\documentclass{article}

\usepackage{iftex}
\usepackage{arxiv}

\usepackage[T1]{fontenc}
\usepackage{graphicx}
\usepackage{xcolor}
\usepackage{hyperref}
\usepackage{amsfonts}
\usepackage{microtype}
\usepackage{xurl}
\ifxetex
    \usepackage{fontspec}
\else
    \usepackage[utf8]{inputenc}
\fi
\begin{document}
\newcommand{\todo}[1]{\textcolor{red}{TODO: #1}}
\renewcommand{\headeright}{\relax}
\renewcommand{\undertitle}{\relax}
\date{}
\title{ProMap: Datasets for Product Mapping in E-commerce}

\author{\href{https://orcid.org/0000-0001-9815-2763}{\includegraphics[scale=0.06]{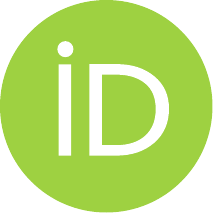}\hspace{1mm}Kateřina Macková} \\
	Charles University, Faculty of Mathematics and Physics\\
    Malostranské náměstí 25\\ 
    Prague 1, 118 00, Czech Republic\\
	\texttt{Katerina.Mackova@mff.cuni.cz} \\
	%% examples of more authors
	\And
	\href{https://orcid.org/0000-0003-1239-1566}{\includegraphics[scale=0.06]{orcid.pdf}\hspace{1mm}Martin Pilát} \\
	Charles University, Faculty of Mathematics and Physics\\
    Malostranské náměstí 25\\ 
    Prague 1, 118 00, Czech Republic\\
	\texttt{Martin.Pilat@mff.cuni.cz} \\ 
}

\maketitle

\begin{abstract}
The goal of product mapping is to decide, whether two listings from two different e-shops describe the same products. Existing datasets of matching and non-matching pairs of products, however, often suffer from incomplete product information or contain only very distant non-matching products. Therefore, while predictive models trained on these datasets achieve good results on them, in practice, they are unusable as they cannot distinguish very similar but non-matching pairs of products. This paper introduces two new datasets for product mapping: ProMapCz consisting of 1,495 Czech product pairs and ProMapEn consisting of 1,555 English product pairs of matching and non-matching products manually scraped from two pairs of e-shops. The datasets contain both images and textual descriptions of the products, including their specifications, making them one of the most complete datasets for product mapping. Additionally, the non-matching products were selected in two phases, creating two types of non-matches -- close non-matches and medium non-matches. Even the medium non-matches are pairs of products that are much more similar than non-matches in other datasets –- for example, they still need to have the same brand and similar name and price. After simple data preprocessing, several machine learning algorithms were trained on these and two the other datasets to demonstrate the complexity and completeness of ProMap datasets. ProMap datasets are presented as a golden standard for further research of product mapping filling the gaps in existing ones.

\end{abstract}


\section{Introduction}

Product mapping or product matching (PM) is the process of matching identical products from different e-shops, where each product can be described by different graphical and textual data. It has an important application in e-commerce as it allows for general marketplace analysis and price comparison among different e-shops. Product mapping is challenging as there is no general identification of products available on all the websites. Therefore, models for measuring similarity of the products need to be trained to identify matching pairs of products based on as much textual and image data describing each product as available. Existing freely available datasets \cite{wdc}, \cite{amazon-walmart}, \cite{prod-map-data-source}, \cite{prod-map-data-source-web} are limited as they often do not provide all data describing each product. Moreover, they consist only of very distant non-matches. Therefore, training the predictive models to distinguish matches and non-matches is very simple. To fill this gap, we create a new group of freely available datasets for product matching by manually scraping selected e-shops and searching for matching and non-matching pairs of products that create a good challenge for further research. In this paper, we present a dataset for Czech product mapping - ProMapCz and a dataset for English product mapping - ProMapEn.

We created these datasets for product mapping by selecting some products from different categories from one e-shop and manually searching for matching pairs in another e-shop. We scraped all available product information including name, price, images, long and short descriptions, specification and source URL. The comparisons were made manually without any type of ID by comparing the available information and then again manually checked. 

The ProMapCz dataset consists of 1,495 matching and non-matching pairs from different categories for training models for Czech product mapping from Alza.cz and Mall.cz e-shops. 34 percent of products are matching, 30 percent are close non-matches, and the rest are medium non-matches. All the available product information is in the Czech language. ProMapEn consists of 1,555 matching and non-matching pairs from different categories for training models for English product mapping from Walmart.com and Amazon.com e-shops. 32.7 percent of products are matching, 32.7 percent are close non-matches, and the rest are medium non-matches. All the available product information is in the English language. Both datasets are available at \url{https://github.com/kackamac/Product-Mapping-Datasets}.
% \url{https://www.dropbox.com/s/t3urznhqtxotydt/Product-Mapping-Datasets.zip?dl=0}
% \footnote{URL will be changed to GitHub for the final version of the paper}.

As a baseline for further research, we preprocessed these datasets and trained several machine-learning models to solve product mapping tasks in English and Czech. We reached the best results with neural network-based models -- F1 scores of 0.777 and 0.706 on ProMapCz and ProMapEn respectively. We also preprocessed and trained the models on two existing datasets -- Amazon-Walmart \cite{amazon-walmart} and Amazon-Google \cite{prod-map-data-source-web,prod-map-data-source} and compared the results to show that the newly proposed datasets are indeed more challenging and provide a good benchmark for product mapping models.

In the rest of the paper, we first discuss existing product matching datasets and their shortcomings in the next section. In Section~\ref{sec:promap}, we describe in detail, how the ProMap datasets were created. Section~\ref{sec:preproc} details the preprocessing performed on these datasets and the following section contains results obtained by a number of different machine learning methods.

\section{Existing Datasets}

We begin with a survey of existing product mapping datasets with emphasis on their deficiency for product mapping tasks to explain the reasons for the creation of new datasets.

\textbf{Web Data Commons (WDC) Dataset}. The WDC Dataset \cite{wdc} is a freely available dataset designed for product mapping. It was the first large dataset that allowed for the training of deep neural network-based models that outperformed traditional symbolic methods on less structured data. WDC Dataset contains 26 million product offers in many languages (out of which 16 million are in English) originating from 79,000 websites (out of which 43,000 are English language sites). The dataset was created automatically using the fact that a lot of e-shops annotate the products with schema.org annotations that contain various IDs. A gold standard dataset was also created by manually verifying 2,200 pairs of offers belonging to four product categories. Altogether with a subset of the training dataset, it is possible to train deep neural network-based approaches for product mapping. Despite the dataset being large enough, it contains only the names of the products. A lot of product characteristics are hidden in descriptions, specifications, and images that are present in almost every e-shop and they are not included in this dataset.

\textbf{Amazon-Walmart}. The Amazon-Walmart dataset~\cite{amazon-walmart} is a collection of 24,583 individual products containing detailed product information such as title, brand, category, price, short description, long description, technical details, image URLs, etc. The product information was collected independently from two e-shops -- Walmart.com and Amazon.com. A deduplication algorithm was performed to obtain 1,154 pairs of matching products between the two data sources creating a dataset for training product mapping models. While this dataset is much more complete than WDC concerning the included information, it contains only very distant non-matches. This is not very useful for product mapping in real life, where detecting close non-matches is very often more important.

\textbf{Amazon-Google}. The Amazon-Google dataset originally published by the Database Group of the University of Leipzig \cite{prod-map-data-source-web,prod-map-data-source} derives from the Amazon.com e-shop and the Google product search service accessible through the Google Base Data API. The dataset contains 1,363 products from Amazon and 3,226 products from Google along with the 1,300 matching record pairs between the two data sources. The selected common product attributes are product name, product description, manufacturer and price. The dataset was created by selecting products from predefined categories on Amazon and based on the selected products, the Google Products dataset was generated by sending queries on the product name. Unfortunately, the dataset does not contain a lot of other useful product information such as specifications or images and does not contain non-matches which need to be generated from matches.

\textbf{Abt-Buy} \textbf{Abt-Buy}. The Abt-Buy dataset also originally published by the Database Group of the University of Leipzig \cite{prod-map-data-source-web,prod-map-data-source} is another dataset for product mapping that contains 1,081 entities from Abt.com and 1,092 entities from Buy.com as well as 1,097 matching product pairs between these two data sources. The selected product attributes are product name, product description and product price. The dataset suffers from the same problem as Amazon-Google: it does not contain all available product information such as specifications or images.

\textbf{Other Datasets}. There exist several datasets for more general task entity matching such as  DBLP-ACM, DBLP-Scholar \cite{prod-map-data-source-web,prod-map-data-source} that contain bibliographic information about books such as title, authors, venue, and year. However, they are not suitable for the product mapping task as they do not store information about products from e-commerce. There is another type of related dataset such as Shopmania \cite{shopmania}, that contains product information organized in a three-level hierarchy comprised of hundreds of categories, or Amazon Review Data \cite{amazonreview} that contains millions of reviews and product metadata (descriptions, category information, price, brand, and image features) from five categories. These datasets contain a lot of product information but they do not contain any product pairs as they are intended for product categorization and not for product mapping.

\textbf{Summary}. We see two deficiencies in the existing datasets -- they either are not complete and do not contain all available product data, or they contain only distant non-matches and, therefore, they are not useful for product mapping in real life, as in practice it is necessary to decide if two very close-looking products are matches or not. In this paper, we therefore created two new datasets having scraped all available product information including images and by selecting not only matching but also two levels of close non-matching products for further model training. 

\section{ProMap Datasets}
\label{sec:promap}

We created our datasets in several stages: first, we selected the URLs of the products on the source website, we let the annotators find the matching URLs, close non-matching and medium non-matching URLs on the target website and we manually verified the correctness of the data to ensure their quality. For ProMapCz the source website is Alza.cz and the target is Mall.cz. For ProMapEn the source is Walmart and the target is Amazon.

\subsection{Selection of the Source Pairs}

We created both product mapping datasets with several different categories. We scraped all data from the source e-shop and we selected the 10 most common product categories intending to cover a wide range of different product categories in ProMapCz. In ProMapEn, we also selected the categories according to the most common products but also regarding the selected categories in ProMapCz to cover similar categories for further language comparisons between Czech and English language. From each category, we randomly sampled 100 products.

\subsection{Searching for Matching and Non-matching Pairs} 

For each source product, we searched for three products on the target website: match, close non-match, and medium non-match. A \emph{match} is the identical product on the target website, the only difference we allowed was the product color. A \emph{close non-match} is defined as the product from the target website which is very similar but not the same as the source product. It was supposed to have the same brand, similar name, similar price (at a maximum 20\% difference from the original price) and almost the same attributes. A \emph{medium non-match} is a product that is different but still has a number of similar attributes -- it is supposed to still have the same brand and similar name and price, but the specifications and prices can be more different than for the close non-matches. If there were multiple suitable candidate products, the annotator was supposed to select one randomly. If the target product was not found, the particular source product was left empty. We prohibited repeating target products for different source products. 

The inclusion of different types of non-matches in this dataset allows us to study product matching models in more detail. This also makes the dataset harder, as even medium non-matches are still products from the same category and with the same brand as opposed to being completely different products. This is more realistic as we do not expect to run product matching on completely different products -- these can be often easily identified, for example, by having a different brand or being in a different category.

\subsection{Manual Check of the Data Quality} 

After the annotators created the datasets, we performed manual control of created data by randomly selecting 20 products in each category and checking
corresponding products to ensure data quality and annotators’ correctness and
to avoid and correct misunderstandings and errors.

\subsection{Product Information Extraction}
After having pairs of URLs of corresponding products from both websites, we automatically scraped all possible product data about all the products from the URLs. Specifically, we scraped names, long text descriptions, short descriptions (often stated immediately under the product names), specifications containing technical information, images and prices. The specific column names and datatypes corresponding to these types of information as they are stored in the dataset are described in Table~\ref{tab:datset-structure}. The column names containing the product information extracted from the source URL are suffixed by 1 at the end of each column name. Similarly, column names containing the product information extracted from the target URL are suffixed by 2.

\subsection{Datasets Description}

Using this process, we obtained the two datasets. The ProMapCz dataset contains 1,409 unique products from Mall.cz and 706 unique products from Alza.cz, these are organized as 1,495 pairs of matching and non-matching products. Out of these 504 are matches, 456 are close non-matches are 535 are medium non-matches. The ProMapEn dataset contains 1,555 unique products from Walmart.com and 751 unique products from Amazon.com. These form 1,555 pairs of products divided into 509 matches, 509 close non-matches and 537 medium non-matches. The distribution of the products, matches and both types of non-matches by category, as well as the list of categories is presented in Table~\ref{tab:categories}. In most categories, we were able to find around 50 matching products, however, some categories were harder than others. For example, in the ProMapCZ dataset in the Laptops category we found only 20 matching products. This is mostly given by the fact that Alza.cz carries much wider range of electronics with a large number of different laptop models than Mall.cz. 

\begin{table}[t]
        \centering
        \caption{\label{tab:datset-structure} Dataset column names with their types.}
        \begin{tabular}{ll|l}
            \hline
            \multicolumn{2}{c|}{\textbf{Name}} & \textbf{Type} \\\hline
              url1 & url2 & URL \\
              name1 & name2 & text   \\ 
              long\_description1 & long\_description2 & text \\
              short\_description1 & short\_description2 & text   \\ 
              specification1 & specification2 & text \\
              images1 & images2 & URLs   \\ 
              price1 & price2 & number \\
              \multicolumn{2}{c|}{match}  & 1/0   \\\hline
        \end{tabular}
\end{table}

\begin{table}[t]
    \centering
     \caption{\label{tab:categories} Number of matching (=), close non-matching (c$\neq$) and medium non-matching (m$\neq$) products in each category of ProMapCz and ProMapEn}
        \begin{tabular}{l|ccc} \hline
            \textbf{ProMapCz Categories} & = & c$\neq$ & m$\neq$\\\hline
            Pet Supplies & 30 & 24 & 30 \\
            Backpacks\&Bags & 35 & 20 & 44 \\
            Hobby\&Garden & 68 & 51 & 50 \\
            Appliances  & 61 & 37 & 46 \\
            Mobile phones & 50 & 29 & 44 \\
            Household Supplies & 46 & 53 & 59 \\
            Laptops & 20 & 50 & 49 \\
            TVs  & 71 & 68 & 71 \\
            Headphones & 65 & 63 & 85 \\
            Fridges  & 58 & 61 &  57 \\\hline
            \multicolumn{4}{c}{~} \\
            \hline
            \textbf{ProMapEn Categories} & = & c$\neq$ & m$\neq$\\\hline
            Pet Supplies & 48 & 30 & 29 \\
            Backpacks\&Accessories  & 43 & 51 & 43 \\
            Patio\&Garden & 47 & 52 & 52 \\
            Kitchen Appliances & 83 & 75 & 81 \\
            Mobile Phones & 45 & 46 & 53 \\
            Home Improvement & 48 & 52 & 52 \\
            Laptops & 40 & 58 & 59 \\    
            Toys & 63 & 40 & 59 \\
            Sports\&Clothes & 39 & 38 & 39 \\
            Health\&Beauty & 48 & 67 & 70\\\hline
        \end{tabular}
\end{table}

\section{Dataset Preprocessing and Similarity Computations}
\label{sec:preproc}

We preprocessed both datasets to train multiple baseline models for the product mapping task. The images and text attributes of each pair of products are preprocessed and converted to numerical vectors and the cosine distance between each pair of corresponding attributes is computed. Additionally, we performed keyword detection in textual attributes and computed the ratio of matching keywords between products for each textual column separately. In this manner, we created a vector of 34 features representing the distances of all attributes between each product pair, see Table \ref{tab:features}. Along with the label whether the pair is matching or not, we used these vectors to train several machine learning models to predict corresponding pairs of products between the two e-shops. The feature extraction process is described in more details below.

\subsection{Image Preprocessing and Similarity Computations}

Each product is represented by several images. We preprocessed all these images to obtain their numerical representation using a perceptual hashing method. Having these hashes, we computed the similarities of images between each product pair to obtain one number representing the products' image similarity.

\textbf{Image preprocessing}. The images of the products often vary in sizes, colors, rotations and centering. They also often contain large white borders and are centered or cropped differently. Therefore, we performed object detection, selected the largest object and cropped the image around it. The process consists of resizing the image into maximal width and height to preserve memory and increase speed. We experimentally set the size to $1024 \times 1024$. We added a white border around the image to allow easier object detection and we converted the image into grayscale. Then, we created a black and white mask of the object and applied canny edge detection and we found contours using methods from the OpenCV\footnote{\url{https://docs.opencv.org/}} library. Finally, we selected the largest object in the picture and cropped the image to its bounding box to obtain the final preprocessed image. In preliminary experiments, we also tried to replace object detection with simpler techniques such cropping of the white borders to the closest edge of the product or by using contour detection, but these did not work that well due to the presence of various logos and other elements added by the e-shops around the images.

\textbf{Image hash creation}. After having the images preprocessed, we created the image hashes preserving the main features and taking much less memory. To this end, we used the Image Perceptual Hashing method~\cite{image-hashes}. We used the implementation from image-hash Node.js library\footnote{\url{https://www.npmjs.com/package/image-hash}}. This technique works by splitting the array of pixels into several blocks and performing kernel operations for finding the main features and creating a hash representing the image and preserving the most important information from it. We used hashes of size 8 with 8 blocks thus creating 64-bit hashes. We selected these parameters experimentally as smaller sizes did not provide enough information to compute the similarity of the images and larger sizes unnecessarily slowed down the run of the algorithm without improving the results.

\textbf{Image hashes similarity computation}. Having a set of image hashes characterizing each product, we compared the hashes of images between corresponding product pairs to obtain overall image similarity. For each image in the source product image set, we computed the Hamming distance of all images in the target product image set and we selected the most similar image. We kept only images having a similarity higher than a given threshold to filter out images that are present in only one of the image sets. The threshold was set experimentally to 0.9 concerning image hashes sizes meaning that images having similarities above 90\% are corresponding to each other. By finding the most similar image and filtering away too-distant images, we obtained the closest image from the second product image set for each image from the first product image set. Finally, we computed the overall similarity of images by summing up all these precomputed similarities. This summing means that products with a greater number of more similar images are more similar. The number obtained in this way is referred to as \textit{hash\_similarity} in the rest of the paper.

\subsection{Text Preprocessing and Similarity Computations}
Each product is represented by several textual attributes: name, long and short description, and specification. We preprocessed all these attributes in the same way to create numerical vectors. We also extracted several main keywords in attributes to create additional vectors. In the end, we computed the cosine distance of corresponding attributes between both products in the given product pair to obtain additional similarity measures.  

\textbf{Text preprocessing}. We preprocessed every attribute containing some text in every product by the same procedure. First, we deleted useless characters from the text such as additional tabs, spaces, brackets etc. Afterwards, we detected the cases of units and values not separated by space (e.g. 15GB or 4") and we separated them using space. We separated the whole text into single words and we lemmatized them using the morphological analyzer Majka~\cite{majka} to avoid problems with declination and conjugation which are quite common in some languages such as Czech. The last step in the preprocessing pipeline is lowercasing the text.

Additionally, we added an \textit{all\_texts} meta-attribute consisting of the concatenation of the name, long and short descriptions, and specification. This allows us to compute features not only for each separate attribute but also for all the texts at once.

\textbf{Text similarity computations}. After having all the texts of each product preprocessed, the similarity of each attribute in each product pair needs to be computed. The final similarity number is computed by converting preprocessed texts into numerical vectors using tf.idf and computing the cosine similarity of both vectors defined as 
    $$\mathrm{similarity}(v_1, v_2) = \frac{v_1\cdot v_2}{|v_1||v_2|},$$
where $v_1 \cdot v_2$ is the dot product of $v_1$ and $v_2$, and $|v_1|$,$|v_2|$ are the lengths of the vectors.

This creates three additional similarity measures: \textit{name\_cos}, \textit{short\_description\_cos}, \textit{long\_description\_cos}. Note that we do not use cosine similarity of the specifications of the products as this field has a different structure and its similarity is computed differently. 

\textbf{Keywords detection and similarity computations}. The textual field can contain a number of words with special meaning, therefore, we decided to perform detection and comparison of such keywords -- IDs, brands, numbers, and descriptive words. The similarity of detected keyword sets in a product pair is computed as the ratio of the same keywords and the number of all keywords for every detected keyword type, i.e as the Jaccard similarity of the two sets defined as $$J(A,B) = \frac{|A \cap B|}{|A \cup B |},$$ where $A$ and $B$ are the two sets of keywords.

The detection of each of the types of keywords is described below:

\begin{enumerate}
    \item \textbf{ID detection} is performed by selecting unique words longer than five characters that are not included in the vocabularies of English and Czech words created based on the ParaCrawl \cite{paracrawl} corpus. ID detection is important as some of the products may have a unique identification that can facilitate the identification of matching products between e-shops. IDs are detected and compared in the name, short description and all text attributes giving us: \textit{name\_id}, \textit{short\_description\_id}, \textit{all\_texts\_id} similarities.

    \item \textbf{Brand detection} is based on our vocabulary created by scraping all brands on the source and target website. It is important as brands in names are another important identification of matching products. Brands are detected and compared in name, short description and all texts attributes, giving us \textit{name\_brand}, \textit{short\_description\_brand}, \textit{all\_texts\_brand} similarities.

    \item \textbf{Numbers detection} is based on detecting numbers in the text and searching for units around them. If no units are found neat the number, the number is detected as a \emph{free number}. It is important as such numbers can contain for example model numbers or other crucial information. Free numbers are detected and compared in every attribute giving us the features:  \textit{name\_numbers}, \textit{short\_description\_num\-bers},  \textit{long\_description\_numbers}, \textit{specification\_text\-\_numbers}, and \textit{all\_texts\_numbers}. 
    \item \textbf{Descriptive words} are a set of the most characterising words for each attribute of the product. These words are selected as the top $k$ words that occur in a maximum of $p$ per cent of documents (in our case among all textual attributes of all products). We experimentally set the $k$ to 50 and $p$ to 50\% to eliminate the most common words such as 'and', 'or' etc. having no important meaning but to preserve words characterising each attribute of the products. Descriptive words are also detected and compared in every attribute, giving us \textit{name\_descriptives}, \textit{short\_description\_descriptives}, \textit{long\_description\_descriptives}, \textit{all\_texts\_descriptives} similarities.

    \item \textbf{Units detection} is based on the extraction of numbers followed by units from each attribute that can specify the product in detail. Units are extracted and compared in all attributes, giving us \textit{name\_units}, \textit{short\_description\_u\-nits}, \textit{long\_description\_units}, \textit{specification\_text\_units}, \textit{all\_texts\_units similarities}.
    
    \item \textbf{Words} are a ratio of the same words taking all words from corresponding attributes of two products. Words are computed only between names, short descriptions and all texts, giving us \textit{name\_words}, \textit{short\_description\_words}, \textit{all\_texts\_words} similarities.

\end{enumerate}

\textbf{All detected keywords comparisons}. We created one list from each detected units, IDs, numbers and brands for each product and we compared the ratio of matching values in those lists between two compared products giving us \textit{all\_units\_list}, \textit{all\_ids\_list}, \textit{all\_numbers\_list}, \textit{all\_brands\_list} similarity numbers allowing to compare detected keywords across all text attributes. We computed the similarity of the detected keyword using the ratio of the same keywords as in other texts.

\textbf{Specification preprocessing}. As the product specification is often in a specific format containing parameter names and their values, we also performed preprocessing of the specification based on the comparisons of these parameters. We extracted numbers followed by units from each attribute and compared them between two products to obtain parameters and their similarity as they can specify the product in detail. As the parameter values can be in different units, we transformed all values to metric units in basic form without any prefixes. We computed the ratio of corresponding parameter names \textit{specification\_key\_matches}, and ratio of corresponding parameter names and values \textit{specification\_key\_value}. While comparing the values, 5\% deviation is allowed to account for inaccuracies during conversions.

\section{Experiments and Results}
After processing text and image attributes, keywords detection and similarity computations, we obtained the overall similarity of two products characterized by a vector containing 34 features. All the features are listed in Table \ref{tab:features}. Along with the \textit{match} label representing either matching or non-matching products, we trained logistic regression (LR), support vector machines (SVM), decision trees (DT), random forests (RF) and neural network (NN) classifiers to predict matching and non-matching pairs.

\begin{table*}[t]
    \small
    \centering
    \caption{\label{tab:features} Precomputed similarity features.}
    \begin{tabular}{llll} 
        \hline
            \multicolumn{4}{c}{\textbf{Feature Name}}\\\hline 
            name\_cos                  & long\_description\_descriptives  & all\_texts\_cos          & all\_numbers\_list           \\
name\_id                   & long\_description\_units         & all\_texts\_brand        & all\_ids\_list               \\
name\_brand                & short\_description\_cos          & all\_texts\_id           & all\_units\_list             \\
name\_numbers              & short\_description\_id           & all\_texts\_numbers      & specification\_text\_numbers \\
name\_descriptives         & short\_description\_brand        & all\_texts\_descriptives & specification\_text\_units   \\
name\_units                & short\_description\_numbers      & all\_texts\_units        & specification\_key           \\
name\_words                & short\_description\_descriptives & all\_texts\_words        & specification\_key\_value    \\
long\_description\_cos     & short\_description\_units        & all\_brands\_list        & hash\_similarity             \\
long\_description\_numbers & short\_description\_words        &                          &                              \\
\hline
    \end{tabular}
\end{table*}

\begin{table}[t]
\centering
\caption{Parameter settings for the grid search.\label{tab:grid}}
\begin{tabular}{l|l|l}
\hline
Model & Parameter & Possible Values \\
\hline
LogisticReg.  & penalty &  l1, l2, elasticnet, none \\
                    & solver & lbfgs, newton-cg, liblinear \\
~&~& sag, saga \\
            ~      & max\_iter & 10, 20, 50, 100, 200, 500 \\                           
\hline
SVM & kernel & linear, poly, rbf, sigmoid \\
                     & degree& 2, 3, 4, 5 \\
                     & max\_iter& 10, 20, 50, 100, 200, 500 \\
\hline
DecisionTree & criterion& gini, entropy \\
             & max\_depth & 5, 10, 15, 20 \\
~& min\_samples\_split & 2, 5, 10, 15, 20 \\
\hline
RandomForest & n\_estimators & 50, 100, 200, 500 \\
             & criterion & gini, entropy \\
~& max\_depth & 5, 10, 20, 50 \\
~& min\_samples\_split & 2, 5, 10, 20 \\
\hline
NeuralNetwork & hidden\_layer\_sizes & (10, 10), (50, 50), (10, 50) \\
             &~&             (10, 10, 10), (50, 50, 50) \\ 
 ~&~&(50, 10, 50), (10, 50, 10) \\
 ~ & activation & relu, logistic, tanh\\
 ~&solver& adam, sgd, lbfgs \\
 ~&learning\_rate & constant, invscaling, adaptive \\
 ~&learning\_rate\_init & 0.01, 0.001, 0.0001 \\
 ~& max\_iter &50, 100, 500 \\
 \hline
\end{tabular}
\end{table}

\textbf{Train-test data preparation}. We split ProMap datasets into train-test data with a ratio of 80:20 giving us 1,196 vectors for training and 299 for testing in ProMapCz and 1,244 vectors for training and 311 for testing in ProMapEn. We release this train-test split to enable the training of other machine learning models and to compare their results in future research.

\textbf{Training and parameters tuning}. We trained linear regression, support vector machines, decision trees, random forests and neural network models with different combinations of values of their hyper-parameters. We performed grid search and random search to find the most suitable hyper-parameters for each model. These hyper-parameter searches use 20 percent of the training set as a validation set. The possible parameter values for grid search are summarized in Table~\ref{tab:grid}. The random search was performed with 100 samples and it uses the same parameter sets for categorical parameters and uses the full range between the minimum and maximum values for the numeric parameters. After the hyper-parameter search, the models are trained with the best hyper-parameters on the combined training and validation sets. We select the model that maximizes the F1 score. The threshold for classification is set such that the F1 score on the training set is maximized.

\subsection{Amazon-Walmart and Amazon-Google Datasets}
\textbf{Amazon-Walmart}. The original Amazon-Walmart dataset contains 22,073 products from Amazon and 2,554 products from Walmart organized as pairs of 1,154 matches and 11.540 non-matches. We merged Amazon and Walmart products using these mappings provided in the dataset and we sampled 1,143 matches and 2,000 non-matches. So that the final dataset has similar distribution of matches and non-matches as the ProMap datasets. The data were split 80:20 into training and testing datasets. The ratio of matches is 36.4\% which approximately corresponds to the ratio of one-third of matches in ProMap datasets. There are 1,640 unique Amazon products and 1,552 unique Walmart products. The Amazon-Walmart dataset contains only name1, short\_description1, long\_description1, price1, id1, name2, short\_description2, long\_description2, price2, id2 and match columns. Image\_hashes and specification columns are not present in this dataset.

\textbf{Amazon-Google}. The Amazon-Google dataset contains 1,363 products from Amazon and 3,326 products from Google with 1,300 matches. We also merged the Amazon and Google products using the provided mappings to obtain a file of 1,300 pairs. We also created 1,935 non-matching pairs by random selection of Google products that do not match any of Amazon's products. We paired those with random products from the whole Amazon file in which all products were doubled to prevent too frequent occurrence of one Amazon product in the final dataset. The newly created Amazon-Google dataset has a total length of 3,235 pairs, of which 2,588 are in the training set and 647 in the testing set to preserve the 80:20 ratio as in ProMap datasets. The ratio of matches is 39.0\% which approximately corresponds to the ratio of one-third of matches in ProMap datasets. There are 1,341 unique Amazon products and 3,326 unique Google products. The created Amazon-Google dataset contains only name1, short\_description1, price1, id1, name2, short\_description2, price2, id2 and match columns. Image\_hashes, long\_description and specification columns are not present in this dataset.

\textbf{Preprocessing and similarity computations} We preprocessed columns in both Amazon-Walmart and Amazon-Google datasets and computed the similarities by the same technique as in ProMap datasets, which gave us the same column of similarities except for specification and image-related ones in the case of Amazon-Walmart and except for specification, long description and image-related ones in Amazon-Google. We publish newly created versions of Amazon-Walmart and Amazon-Google datasets including their train-test split and precomputed similarities along with our ProMap datasets.  

\textbf{Training and evaluation}. We have performed the same training and evaluation techniques on Amazon-Walmart and Amazon-Google as on the ProMap dataset including the parameters search. Moreover, we tested the transfer learning capabilities of the models by selecting the best model for every dataset and evaluating it in all the other datasets. As each of the datasets contain different attributes, we filled the values of the missing attributes with zeroes for models that were trained on datasets containing them and we removed the additional attributes for models trained on datasets not containing them.

\subsection{Results}

\begin{table*}[t]
        \small
    \centering
    \caption{\label{tab:results} Comparison of several machine learning methods trained on ProMapCz, ProMapEn, Amazon-Walmart and Amazon-Google datasets evalueated on their test datasets. Results are from the models with the best parameters from random and grid searches. The model name contains the base model and the type of hyper-parameter search used.}
    \begin{tabular}{l|ccc|ccc|ccc|ccc}
        \hline
         ~ & \multicolumn{3}{|c|}{\textbf{ProMapCz}} & \multicolumn{3}{|c}{\textbf{ProMapEn}} & \multicolumn{3}{|c|}{\textbf{Am-Walmart}} & \multicolumn{3}{|c}{\textbf{Am-Google}}
         \\\hline
         \textbf{Model} & \textbf{F1} & \textbf{Prec} & \textbf{Rec} & \textbf{F1} & \textbf{Prec} & \textbf{Rec} & \textbf{F1} & \textbf{Prec} & \textbf{Rec} & \textbf{F1} & \textbf{Prec} & \textbf{Rec} \\\hline 
         LR-Rand  & 0.754 & 0.774 & 0.735 & 0.692 & 0.683 & 0.703 & 0.921 & 0.917 & 0.925 & 0.990 & 0.989 & 0.992 \\
         LR-Grid  & 0.750 & 0.766 & 0.735 & 0.701 & 0.731 & 0.673 & 0.923 & 0.921 & 0.925 & 0.990 & 0.989 & 0.992 \\
         SVM-Rand  & 0.733 & 0.659 & 0.827 & 0.631 & 0.627 & 0.634 & 0.916 & 0.895 & 0.939 & 0.985 & 0.992 & 0.977 \\
         SVM-Grid  & 0.739 & 0.714 & 0.765 & 0.664 & 0.628 & 0.703 & 0.927 & 0.908 & 0.947 & 0.985 & 0.996 & 0.973 \\
         DT-Rand  & 0.737 & 0.761 & 0.714 & 0.626 & 0.535 & 0.752 & 0.896 & 0.902 & 0.890 & 0.983 & 0.981 & 0.985 \\
         DT-Grid  & 0.732 & 0.740 & 0.724 & 0.626 & 0.535 & 0.752 & 0.901 & 0.926 & 0.877 & 0.987 & 0.989 & 0.985 \\
         RF-Rand  & 0.760 & 0.890 & 0.663 & 0.647 & 0.573 & 0.743 & 0.918 & 0.924 & 0.912 & 0.982 & 0.996 & 0.969 \\
         RF-Grid  & 0.759 & 0.868 & 0.673 & 0.658 & 0.574 & 0.772 & 0.924 & 0.911 & 0.939 & 0.985 & 0.992 & 0.977 \\
         NN-Rand  & 0.777 & 0.790 & 0.765 & 0.690 & 0.697 & 0.683 & 0.912 & 0.872 & 0.956 & 0.990 & 0.996 & 0.985 \\
         NN-Grid  & 0.762 & 0.700 & 0.836 & 0.706 & 0.710 & 0.703 & 0.928 & 0.901 & 0.956 & 0.990 & 1.000 & 0.981 \\
         \hline
    \end{tabular}
\end{table*}

\begin{table*}[t]
        \small
    \centering
    \caption{\label{tab:results3} Comparison of the best neural network models. One model was trained on each dataset and evaluated on all datasets to show the difficulty of ProMap datasets.}
    \begin{tabular}{c|ccc|ccc|ccc|ccc} 
        \hline 
        \textbf{}   &  \multicolumn{12}{|c}{\textbf{Train data}} \\\hline
        \textbf{}   & \multicolumn{3}{|c|}{\textbf{ProMapCz}} & \multicolumn{3}{|c|}{\textbf{ProMapEn}} & \multicolumn{3}{|c|}{\textbf{AmWa}} & \multicolumn{3}{|c}{\textbf{AmGo}}
         \\\hline
         \textbf{Test data}   & \textbf{F1} & \textbf{Prec} & \textbf{Rec} & \textbf{F1} & \textbf{Prec} & \textbf{Rec} & \textbf{F1} & \textbf{Prec} & \textbf{Rec} & \textbf{F1} & \textbf{Prec} & \textbf{Rec} \\\hline 
         \textbf{ProMapCz} & 0.762 & 0.700 & 0.836 & 0.710 & 0.647 & 0.786 & 0.678 & 0.569 & 0.837 & 0.582 & 0.438 & 0.867 \\
         \textbf{ProMapEn} & 0.631 & 0.619 & 0.644 & 0.706 & 0.710 & 0.703 & 0.618 & 0.612 & 0.624 & 0.589 & 0.523 & 0.673 \\
         \textbf{AmWa} &     0.859 & 0.863 & 0.855 & 0.915 & 0.909 & 0.921 & 0.928 & 0.901 & 0.956 & 0.990 & 1.000 & 0.981 \\
         \textbf{AmGo} &     0.968 & 0.962 & 0.973 & 0.983 & 0.988 & 0.977 & 0.899 & 0.868 & 0.931 & 0.990 & 1.000 & 0.981 \\
                                                      
        \hline
    \end{tabular}
\end{table*}

The best results were obtained by neural network-based models closely followed by logistic regression and random forests. These results are the same across all datasets confirming the huge potential in the neural network-based models, see Table~\ref{tab:results}. It seems that neural networks tend to have the most balanced precision and recall, while the random forests can have larger differences between these values. Support vector machines seem to be inappropriate for such predictions in general. Models trained on ProMapCz have 6-10 percentage points better results than models trained on ProMapEn which can be caused by several reasons such as language differences, differences in the complexities of the English and Czech dataset, selection of products, or sparsity of the data.

\begin{table*}[t]
    \centering
    \caption{\label{tab:nonmatches_comparison} Evaluation of the selected best neural network models on modified test dataset to compare differences in difficulties of close and medium non-matches in ProMapCz and ProMapen.}
    \begin{tabular}{c|c|ccc}
        \hline
         \textbf{Test dataset version} & \textbf{Dataset length} & \textbf{F1} & \textbf{Prec} & \textbf{Rec}  \\\hline
            ProMapCz (matches+close nonmatches)  & 187 & 0.806 & 0.774  & 0.837 \\
            ProMapCz (matches+medium nonmatches) & 210 & 0.859 & 0.881  & 0.837 \\
            ProMapEn (matches+close nonmatches)  & 195 & 0.743 & 0.789  & 0.703 \\
            ProMapEn (matches+medium nonmatches)  & 217 & 0.780 & 0.877  & 0.703 \\
    \end{tabular}
\end{table*}

\textbf{Complexity of the datasets}. Overall results of models for Amazon-Walmart and Amazon-Google datasets are much higher than for ProMap datasets confirming that ProMap datasets are more challenging. This hypothesis is also confirmed by evaluating the best model for every dataset on all other datasets, see Table~\ref{tab:results3}. The difference between the complexity of Amazon-Walmart\&Amazon-Goople and ProMapCZ\&ProMapEn is significant and we can observe them on every model independent of the training datasets.

\textbf{Grid Search versus Random Search}. There does not seem to be a significant difference between the results found by the grid search and the random search. 

\textbf{Close non-matches versus medium non-matches}. In order to evaluate the effect of the close and medium non-matches, we have selected the two best neural network models (NN-Grid): one trained on ProMapCz and the other trained on ProMapEn. We have evaluated them on four test datasets, two created by removing medium non-matches and two by removing close non-matches from the original ProMapCz and ProMapEn test datasets respectively. We used them to compare the differences in complexity between both types of non-matches. The results prove that close non-matches are more difficult for both models to distinguish from matches than medium-non-matches -- confirming the increased complexity of both the ProMap datasets, see Table \ref{tab:nonmatches_comparison}

\textbf{Czech versus English}. Models trained on the Czech dataset consistently achieve better results than models trained on the English dataset indicating that the English data are more difficult than the Czech data. There could be several reasons for that, such as fewer IDs in the product descriptions in the English e-shops, different selection of products and categories or the need for further finetuning of preprocessing. More detailed analysis of these differences is left as a future work.

\section{Conclusion}

We created two datasets for the product mapping task by manual matching and automated extraction of all possible attributes characterizing products from different categories from two different e-shops in the Czech and English languages. These datasets categorize the non-matches into two categories of close and medium non-matches, making a more detailed analysis of product matching models viable. Additionally, even the medium non-matches are pairs of much more similar products than non-matches in existing product mapping datasets, making the new datasets more challenging for product mapping models.

We performed text and image preprocessing and we trained several machine learning models to predict matching and non-matching pairs of products between two e-shops. We also preprocessed two existing datasets for the same problem and trained several machine learning models on those to prove the increased complexity of our datasets. We present the datasets and models as the baseline for further extensions and improvements in the product mapping task. 

In our further research, we will focus on the extension of the datasets. We plan to create a version of English datasets comparing the same products among several e-shops. We want to focus on the improvement of data preprocessing and filtering, and more powerful feature analysis and key information extraction. Regarding the models, we plan to examine other machine learning methods including those based on deep neural networks.

\section*{Acknowledgements} 
This research was partially supported by SVV project number 260 698.

\sloppy
\bibliographystyle{plainurl}
\bibliography{biblio}
\end{document}